\documentclass[11pt]{article}

\usepackage[final]{acl}

\usepackage{times}
\usepackage{latexsym}

\usepackage[T1]{fontenc}
\usepackage[utf8]{inputenc}

\usepackage{microtype}

\usepackage{inconsolata}

\usepackage{graphicx}

\usepackage{enumitem}

\usepackage{subcaption}
\usepackage{booktabs} % for professional tables

\usepackage{xurl}      % allows line breaks in long URLs
\usepackage{hyperref}
\usepackage{multirow}
\usepackage{array}
\usepackage{makecell}
\usepackage{placeins}
\usepackage{afterpage}
\usepackage{multicol}
\usepackage{enumitem}

\usepackage{algorithm}
\usepackage{algpseudocode}

\usepackage[colorinlistoftodos,prependcaption,textsize=tiny]{todonotes}

\usepackage{amsmath}
\usepackage{amssymb}
\usepackage{mathtools}
\usepackage{amsthm}
\usepackage{bbm}
\usepackage[most]{tcolorbox}
\usepackage{float}
\usepackage{ragged2e}
\usepackage{capt-of}
\usepackage{tabularx}
\usepackage[capitalize,noabbrev]{cleveref}

\usepackage{stfloats}

\theoremstyle{plain}

\theoremstyle{definition}

\theoremstyle{remark}

\usepackage[textsize=tiny]{todonotes}

\newcounter{datasetex}

\newcommand{\smallqed}{\hspace{1px}\rule{1.1mm}{1.1mm}}

\definecolor{examplebg}{RGB}{245, 250, 254}  % Light blue background
\definecolor{exampleborder}{HTML}{1D4688}  % Light blue border
\definecolor{exampletitle}{HTML}{000000}  % Light blue background
\newcommand{\titlecite}[1]{{\normalfont\normalcolor\cite{#1}}}

\newtcolorbox{datasetexample}[2][]{%
  enhanced,
  breakable,
  colback=examplebg,
  colframe=quoteborder,
  arc=0mm,
  sharp corners,
  left=10pt,
  right=10pt,
  leftrule=3pt,
  rightrule=0pt,
  bottomrule=0pt,
  toprule=0pt,
  top=7pt,
  bottom=7pt,
  before skip=10pt,
  after skip=10pt,
  parbox=false,
  before upper={%
      \begingroup  
      \color{exampletitle}\textbf{#2}\par\vspace{3pt}%
      \endgroup
      },
  fontupper=\small\justifying,
  #1
}

\definecolor{quotebg}{HTML}{FFFBED}     
\definecolor{quoteborder}{HTML}{eec848}   
\newtcolorbox{quotedef}{
    enhanced,
    colback=quotebg,
    colframe=quoteborder,
    arc=0mm,
    boxrule=0pt,
    leftrule=3pt,
    rightrule=0pt,
    toprule=0pt,
    bottomrule=0pt,
    left=11pt,
    right=11pt,
    top=7pt,
    bottom=7pt,
    fontupper=\small\justifying,
    sharp corners,
    before skip=10pt,
    after skip=10pt,
    parbox=false
}

\definecolor{hlcolor}{HTML}{FFFCED} 

\newcommand{\customhighlight}[1]{%
  \begingroup\setlength{\fboxsep}{1pt}\colorbox{hlcolor}{#1}\endgroup%
}

\definecolor{tagthink}{HTML}{3962a4}
\definecolor{tagtool}{HTML}{ba9100}
\definecolor{tagrollout}{HTML}{ba9100}
\definecolor{taganswer}{HTML}{3962a4}

\newcommand{\ewmThinkTag}{\textcolor{tagthink}{\texttt{<think>...\allowbreak</think>}}}
\newcommand{\ewmToolTag}{%
\texttt{%
\textcolor{tagtool}{<tool\_call>}%
\allowbreak
\textcolor{tagthink}{\{``name'': ``world\_module'', ``query'': q\}}%
\allowbreak
\textcolor{tagtool}{</tool\_call>}%
}%
}
\newcommand{\ewmRolloutTag}{\textcolor{tagrollout}{\texttt{<visual\_rollout>...\allowbreak</visual\_rollout>}}}
\newcommand{\ewmAnswerTag}{\textcolor{taganswer}{\texttt{<answer>...\allowbreak</answer>}}}

\newcommand{\ewmThinkOpen}{\textcolor{tagthink}{\texttt{<think>}}}
\newcommand{\ewmToolOpen}{\textcolor{tagtool}{\texttt{<tool\_call>}}}
\newcommand{\ewmRolloutOpen}{\textcolor{tagrollout}{\texttt{<visual\_rollout>}}}
\newcommand{\ewmAnswerOpen}{\textcolor{taganswer}{\texttt{<answer>}}}

\newcommand{\algmid}{\fontsize{10.5pt}{12pt}\selectfont}

\title{Einstein World Models}

\author{
\textbf{Munachiso Samuel Nwadike$^{1,2}$, Zangir Iklassov$^{1}$, Ali Mekky$^{1}$}\\ 
\textbf{Zayd M. Kawakibi Zuhri$^{1}$, Kentaro Inui$^{1,2,3}$} \\ 
$^{1}$MBZUAI \quad
$^{2}$RIKEN AIP, Japan \quad
$^{3}$Tohoku University \\
\texttt{\footnotesize munachiso.nwadike@mbzuai.ac.ae}
}
\begin{document}
\maketitle

\begin{abstract}
\vspace{-1.5em}
Does \textit{intelligence} require the ability to reason about phenomena beyond direct experience? It is natural to suspect that some complex thought cannot be captured through language alone. However, of particular concern to this work, is whether \textit{visualising} counterfactual events can \textit{complement language} as a mechanism for complex thought. We ask whether LLMs can be trained to utilise such visualisation mechanisms, in a way that benefits their reasoning abilities. Motivated by this question, we propose \textit{Einstein World Models}. EWMs are a blueprint for LLM-based reasoning systems that place visual-temporal rollouts inside the reasoning trace, allowing them to reason in ways that text alone may not support well. In an EWM, the LLM calls a \textit{world-module} (not to be confused with a world model), to produce short rollouts of scenes under consideration. The returned rollout is treated not as the answer, but as an inspectable hypothesis that can support later reasoning. Einstein World Models extend the capability of LLMs for tool calling (such as web search or code execution), into the domain of visual thought experiments. 
\end{abstract}

% For suitable problems, EWMs outline a path toward LLMs that can perform visual scientific thought experiments. 
% \end{abstract} %% SUBTLEY REDUCE USING THE WORD COT IN THE PAPER. KINDA OVERHYPED.

%%%%%%%%%%%%%%%%%%%%%%%%%%%%%%%%%%%%%%%%%%%%%%%%%%%%%%%%%%%%%%%%%%%%%%%%%%%%%%%%%%%%%%%%%%%%%%%%%%%%%%%%%%%%%%%%%%%%

% \setcounter{section}{-1}
% \section{Disclaimer}
\section*{Prologue}
\label{sec:prologue}

\begin{figure*}[b]
  \centering
  % \vspace{-20em}
  \includegraphics[width=1.0\textwidth]{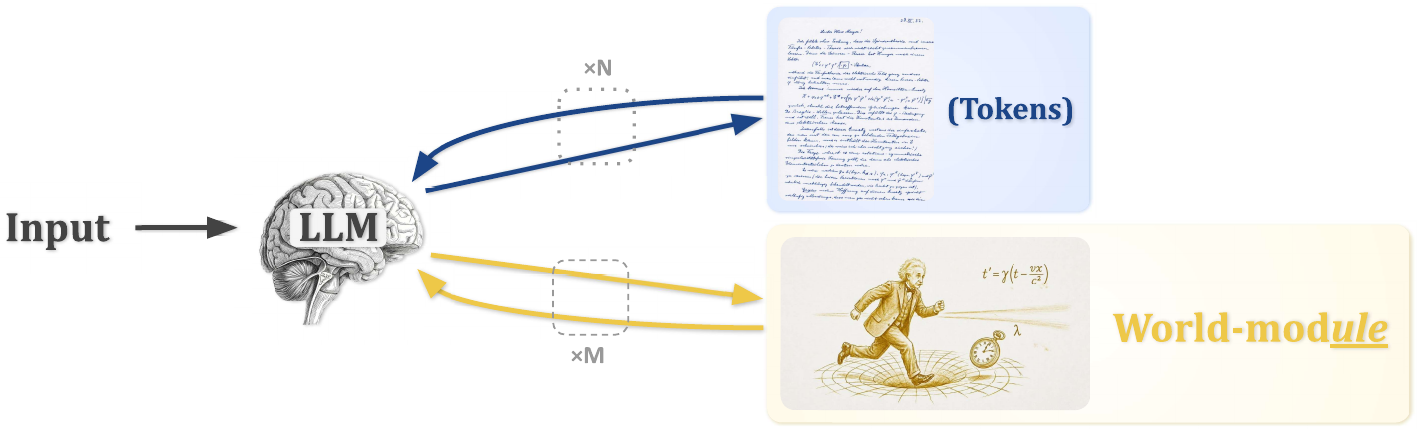}
  % \caption{\customhighlight{\textit{Einstein World Models} (proposed)} output reasoning traces, where \textit{world-module} outputs are one component. The LLM remains the reasoner, while a renderer, simulator, or hybrid world model supplies a callable \customhighlight{world-module}, whose output is treated as an inspectable hypothesis rather than as the answer.} 
  \caption{\customhighlight{\textit{Einstein World Models} (proposed)} build upon traditional LLM reasoning traces. However, in addition to generating tokens across \(N\) autoregressive steps, the model may, at a sparse set of \(M\) intermediate steps, invoke a callable \customhighlight{world-module}. The returned visual-temporal rollout becomes part of the trace as an inspectable hypothesis.}
  \label{fig:ewm_overview} 
\end{figure*}

% Should I say 
% "In the tradition of agenda-setting work on future AI systems and world models, such as \citet{lecun2022path},"

\textit{A call for datasets becomes meaningful after the desirable capability has been specified. This culminates in Section \ref{sec:call_for_datasets}. The predominance of this work therefore aims to motivate a learnable format for visual thought experimentation, proposing the architecture and training objectives necessitating their training data. This work may therefore be understood as an operationalisation of a promising capability whose data requirements are, per this moment, a conceivably fertile frontier.}

%%%%%%%%%%%%%%%%%%%%%%%%%%%%%%%%%%%%%%%%%%%%%%%%%%%%%%%%%%%%%%%%%%%%%%%%%%%%%%%%%%%%%%%%%%%%%%%%%%%%%%%%%%%%%%%%%%%%
\section{Introduction}
\label{sec:introduction}

Scientific invention often makes abstract ideas tractable by turning them into \textit{thought experiments}. In a thought experiment, we visualise a scene, let it unfold, and notice what changes.

Einstein’s recollection of special relativity begins with precisely such a thought experiment, later popularised through the image of Bern’s famous clock tower. In his \textit{Autobiographical Notes} \cite{einstein1949autobiographical}, he recalls imagining what it would be like to chase a beam of light. If he could accelerate until he matched the speed of the beam, would it eventually appear to hang motionless beside him? In those notes, Einstein writes that such a stationary light wave seemed impossible both empirically and according to Maxwell’s equations. His simple thought experiment had \textit{transformed an abstract tension} between electrodynamics and intuitions about motion, into a concrete scene upon which reasoning could be built.

\indent It is precisely in this spirit that ``Einstein World Models'' invoke ``\textit{\underline{E}instein}''. 
Vitally, the ``E'' in EWM also carries a dual reading, overloaded to mean ``\textit{\underline{E}xternalised}''. Externalisation brings the thought experiment into view as an inspectable reasoning trace component, and therefore, as a measurable one \cite{nwadike2026measuringaireasoningguide}, as per chain-of-thought \cite{wei2022chain}. LLMs can already search the web when they lack reliable facts \citep{nakano2021webgpt}, run code when numerical calculation is needed \citep{gao2023pal}, and call external tools when a task is easier to act out than to solve from tokens alone \citep{schick2023toolformer,yao2023react,patil2024gorilla}. What we propose is to render LLMs capable of imagining a scene, when quantitatively beneficial.\\

\indent Hadamard \cite{hadamard1945psychology}, in his inquiry into the inner mental processes behind mathematical invention, recorded Einstein's self-description of the thought experiment process in strikingly imagistic terms:
\begin{figure}[H]
\centering
\vspace{-0.3em}
\begin{minipage}{1.0\linewidth}
\begin{quotedef}
{\footnotesize\raggedleft
--- \textit{Professor Albert Einstein} \cite{hadamard1945psychology}\\}
\vspace{0.5em}
\textit{``The psychical entities which seem to serve as elements in [my] thought are certain signs and more or less clear {\underline{images}} which can be `voluntarily' reproduced and combined.}

\vspace{0.5em}

\textit{\ldots{ }The above mentioned elements are, in my case, of \underline{visual} and some of muscular type. \underline{Conventional words} or other signs have to be sought for laboriously only in a \underline{secondary stage}, when the mentioned associative play is sufficiently established and can be
reproduced at will.''}
\end{quotedef}
\end{minipage}
\vspace{-0.5em}
\end{figure} In a thought experiment, the combination of visuals, albeit subject to meaning, takes precedence, while symbols and words play an augmentative role. Although LLMs already excel with words and symbols in chain-of-thought, their visual-temporal reasoning faculties are yet to capture this dynamic (kindly see Section~\ref{sec:related_work}). Thus, in an effort towards precisely this dynamic, prior work has popularised world models \cite{lecun2022path}. 

These works often characterise world modelling as either passive visual prediction from experience \cite{bardes2024revisiting,garrido2025intuitivephysics}, or action-conditioned prediction for embodied agents \cite{murlabadia2026vjepa21,nam2026causaljepa,maes2026leworldmodel}. However, if world models are treated mainly as high-fidelity simulators of futures tethered either to observed states or to chosen actions, then thought experiments become correspondingly limited either to experience, or to intervention (see Supplementary Notes, Part~\ref{appendix:relation_to_wms}). Einstein World Models reconsider this correspondence. Much like Einstein could visualise riding beside a beam of light without first having to ride one, a language model should be able to benefit from imagining how a described scene could unfold, without necessarily acting within it.

A decisive element in \textit{Einstein World Models} is the choice of a \textit{world-module} capable of producing useful video thought experiment rollouts, as discussed in Section~\ref{sec:world_module_selection}. Equally central is the ability to integrate these rollouts back into the LLM's reasoning trace, as discussed in Section~\ref{sec:ewms}.

Our analysis of this element forms one of three contributions: First, we propose Einstein World Models as a mechanism for selective visual-temporal thought experiments instantiated by tool-use behavior in LLM reasoning. Second, we distinguish Einstein World Models as reasoning systems, from the world-modules they call, treating generated rollouts as inspectable intermediate artifacts. Third, we offer dataset and training recommendations for training LLMs into Einstein World Model reasoners, capable of utilising visual-temporal thought-experiment traces.

%%%%%%%%%%%%%%%%%%%%%%%%%%%%%%%%%%%%%%%%%%%%%%%%%%%%%%%%%%%%%%%%%%%%%%%%%%%%%%%%%%%%%%%%%%%%%%%%%%%%%%%%%%%%%%%%%%%%%%%%%%%%%%%%%%%%%%%%%%%%%%%%%%%%%%%%%%%%%%%%%%%%%%%%%%%%%%%%%%%%%%%%%%%%%%%%%%%%%%%%%%%%%%%%%%%%%%%%%%%%%%%%%%%%%%%%
% \newpage

\begin{figure*}[!t]
\centering
\includegraphics[width=0.905\textwidth]{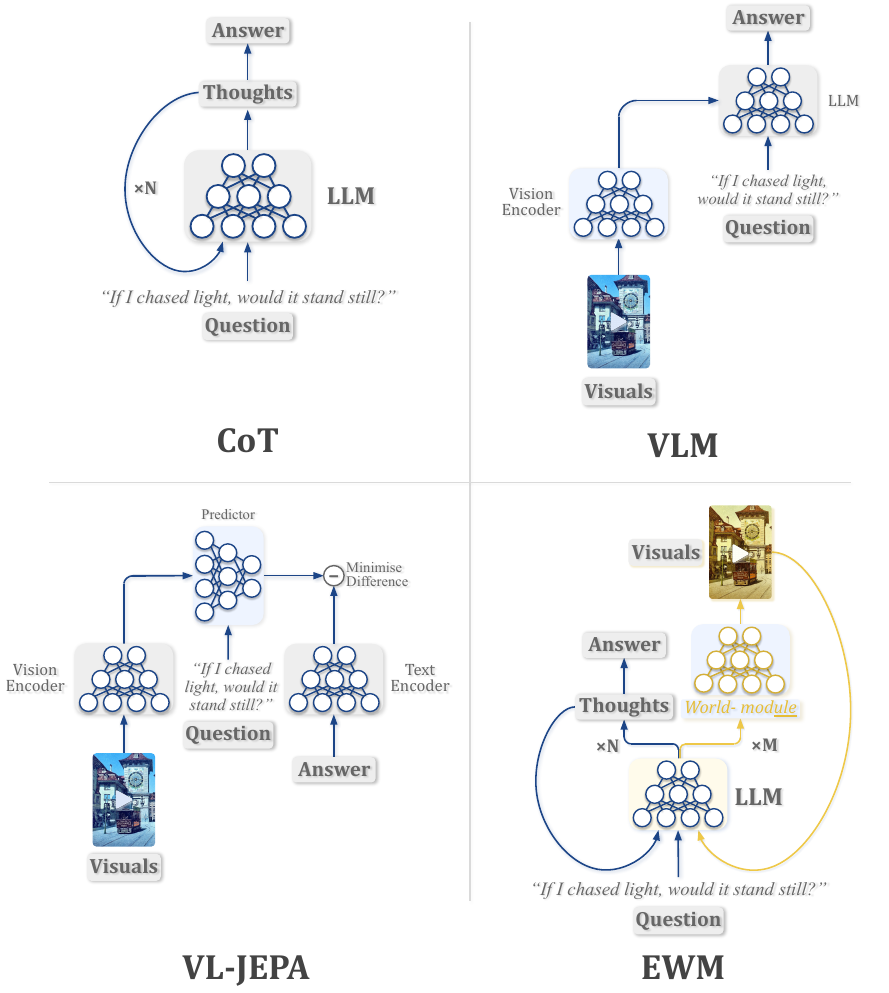}
\caption{
\textit{If Einstein were travelling away from the Bern clock tower at the speed of light, would the clock's hands appear frozen in time?}
This figure compares how different reasoning paradigms would support such a thought experiment.
\textit{Top left:} In CoT, the LLM reasons only through \(N\) autoregressive text-generation steps.
\textit{Top right:} In VLMs, visual information can condition generation, but the relevant visuals are externally provided, rather than visualised during the reasoning trace.
\textit{Bottom left:} In VL-JEPA-style world models, predictive visual representations are learned from observed visual inputs, but the visual prediction itself is not selectively invoked as an intermediate reasoning artifact.
\textit{Bottom right:} In \textit{Einstein World Models}, by contrast, the LLM (the reasoner) preserves autoregressive text generation while taking responsibility for deciding when to invoke a world-module \(M\) times, how to query it, and how to incorporate each returned rollout into subsequent reasoning.
Each rollout therefore enters the trace as a visible thought experiment rather than as a pre-given input or final answer.
}
\label{fig:comparison} 
\end{figure*}

%%%%%%%%%%%%%%%%%%%%%%%%%%%%%%%%%%%%%%%%%%%%%%%%%%%%%%%%%%%%%%%%%%%%%%%%%%%%%%%%%%%%%%%%%%%%%%%%%%%%%%%%%%%%%%%%%%%%%%%%%%%%%%%%%%%%%%%%%%%%%%%%%%%%%%%%%%%%%%%%%%%%%%%%%%%%%%%%%%%%%%%%%%%%%%%%%%%%%%%%%%%%%%%%%%%%%%%%%%%%%%%%%%%%%%%%

\section{Einstein World Models\smallqed}
\label{sec:ewms} 

\subsection{\textit{Overview}}
\label{sec:overview}

The objectives of \textit{Einstein World Models} are twofold. First, we wish to imbue language models with the ability to construct visualisations when answering questions that require physical intuition or scene-level visualisation. Second, we wish to obtain a window into those visualisations. An EWM rollout, as illustrated in Figure~\ref{fig:ewm_overview}, is in effect, a visible hypothesis about how a described scene might unfold.

Let \(\mathcal{T}\) denote a reasoning trace, and let \(N\) denote the number of autoregressive text-generation steps in that trace. A standard CoT trace uses these \(N\) steps to generate text alone. An Einstein World Model also generates text autoregressively, but at a sparse set of \(M\) intermediate generation steps, queries a world-module. Each call to the world-module produces a short video sequence, which is then returned to the reasoner and used to condition subsequent generation. Figure~\ref{fig:comparison} contrasts this structure with CoT, VLMs, and VL-JEPA-style world models.

For problems in which Einstein World Models are useful, we expect that 

\vspace{-0.4em}
\begin{equation*}
0 < M \ll N-1,
\end{equation*}
% \vspace{-1.4em}

as the purpose of a rollout is not to replace the chain-of-thought, but to externalise a visualised scene, at moments where doing so facilitates subsequent reasoning. 

Once returned, the rollout becomes part of the reasoning trace. The choice of prompt used to query the world-module is therefore part of the reasoning problem as well. As with web search, the query helps determine what information is returned. In \textit{Einstein World Models}, the reasoner must learn not only when to visualise, but how to query for the visualised scene that will support later reasoning.

%%%%%%%%%%%%%%%%%%%%%%%%%%%%%%%%%%%%%%%%%%%%%%%%%%%%%%%%%%%%%%%%%%%%%%%%%%%%%%%%%%%%%%%%%%%%%%%%%%%%%%%%%%%%%%%%%%%%%%%%%%%%%%%%

\subsection{\textit{Rollouts as Inspectable Hypotheses}}
\label{sec:inspectability}

Externalised thought makes latent assumptions visible \citep{nwadike2026measuringaireasoningguide}. \textit{Einstein World Models} extend this principle from language to visual-temporal reasoning. Instead of leaving the model's visualised scene implicit, the system renders that scene as an examinable video artifact, for instance as a sequence of frames. 

Because this rollout is externalised, it can be shared and studied \textit{without access to model weights or hidden activations}. In this sense, \textit{Einstein World Models} turn an otherwise private visualised episode into a public object of analysis, even when the underlying model is not itself open to inspection.

Key here is the distinction between the \textit{inspectability} of the world-module's rollout, and the physical \textit{plausibility} of the rollout. A rollout is not considered more plausible solely because it is inspectable. Furthermore, the visual-temporal rollout need not begin from physically ordinary premises in order to be informative. Einstein's own light-chasing scene, for example, was not valuable because it was itself a realisable experiment. To the contrary, it was \textit{conspicuously counterfactual}. However, it was valuable because it made a counterintuitive possibility precise enough to reason about. Similarly, we neither treat the rendered visibility of an EWM rollout as evidence of its plausibility, nor its imperfections as evidence that it is uninformative. What matters is whether the hypothesis it exposes can be inspected, tested, and improved.

%%%%%%%%%%%%%%%%%%%%%%%%%%%%%%%%%%%%%%%%%%%%%%%%%%%%%%%%%%%%%%%%%%%%%%%%%%%%%%%%%%%%%%%%%%%%%%%%%%%%%%%%%%%%%%%%%%%%%%%%%%%%%%%%

\subsection{Inference}
\label{sec:inference}

At inference time, an \textit{Einstein World Model} is specified by two core components
\[
\pi_\theta \quad\text{and}\quad \mathcal{W}.
\]
Here, \(\pi_\theta\) denotes the \textit{Einstein reasoner}, a trainable LLM policy parameterised by \(\theta\), and \(\mathcal{W}\) denotes the world-module. \(\pi_\theta\) generates a reasoning trace and may query \(\mathcal{W}\) for visual-temporal rollouts. If queried, \(\mathcal{W}\) generates these video rollouts, and returns them to \(\pi_\theta\) for further autoregression.

For a text-only input problem \(x\), inference constructs a thought trace with the initialisation
\[
\mathcal{T}_0 = x,
\]
such that \(\mathcal{T}_t\) then denotes the partial trace after step \(t\). At each step, the Einstein reasoner defines a conditional distribution over the next model-generated segment,
\[
\pi_\theta(\cdot \mid \mathcal{T}_t).
\]
A generated segment is either a non-tool segment \(s_t\), such as language reasoning or a final answer, or a world-module query segment \(q_t\). If a query \(q_t\) is generated, the world-module returns a visual-temporal rollout
\[
v_t \sim \mathcal{W}(q_t).
\]
The trace update is therefore
\begin{equation}
\mathcal{T}_{t+1}
=
\mathcal{T}_t
\oplus
\begin{cases}
s_t,
& \text{if } \mathcal{W} \text{ is not queried},\\
[q_t,v_t],
& \text{if } \mathcal{W} \text{ is queried}.
\end{cases}
\end{equation}
Thus, ordinary text segments are appended directly to the trace, while world-module calls append the reasoner-generated query, concatenated with the returned rollout observation. If the generated segment is a final answer, inference terminates and returns that answer.

In implementation, these trace segments can be serialised with special tags, formatted similarly to recent RL-based tool-use systems for search and agentic tool interaction \citep{jin2025searchr1,singh2025artist}. An EWM trace may use \ewmThinkTag{} for language reasoning, \ewmToolTag{} for a world-module query, \ewmRolloutTag{} for the returned visual-temporal rollout, and \ewmAnswerTag{} for the final answer.

The \ewmRolloutOpen{} segment is returned by \(\mathcal{W}\). In practice, \(\mathcal{W}\) may render a short sequence of frames, which can be encoded into visual tokens for incorporation into the reasoning trace. The rendered frames remain available for inspection, while the visual tokens provide the representation consumed by the reasoner.

%%%%%%%%%%%%%%%%%%%%%%%%%%%%%%%%%%%%%%%%%%%%%%%%%%%%%%%%%%%%%%%%%%%%%%%%%%%%%%%%%%%%%%%%%%%%%%%%%%%%%%%%%%%%%%%%%%%%%%%%%%%%%%%%

\subsection{{Training}}
\label{sec:training}

At the initial stage, the reasoner undergoes supervised fine-tuning on EWM trace formats with standard next-token cross-entropy, masking returned rollout observations from the loss as detailed in the Supplementary Notes, part~\ref{appendix:sft}.

Consistent with standard practice in recent reasoning-model training \citep{guo2025deepseekr1}, we then recommend following SFT with RLVR-style training over complete EWM trajectories, since target tasks provide verifiable final answers even when intermediate visual thought experiments remain unlabeled. RL-based tool use already provides a standard methodology for training LLMs to call fixed external systems, such as search engines, and incorporate their returned outputs \citep{jin2025searchr1,qian2025toolrl,singh2025artist}. 
 
Formally, let \(\mathcal{D}=\{(x_i,y_i^\star)\}_{i=1}^{n}\)
be a dataset of text-only problems \(x_i\) with verifiable final answers \(y_i^\star\). A sampled EWM trajectory is a completed thought trace \(\mathcal{T}\) containing all language-reasoning segments, world-module queries, returned rollouts, and the final answer \(\hat{y}\).

Let \(r(\hat{y},y^\star)\) denote the verifier reward for the final answer. For exact-answer tasks, this may simply be \(r(\hat{y},y^\star) = \mathbf{1}[\hat{y}=y^\star].\)

Since world-module calls may be computationally expensive and should be used selectively, we define an EWM reward that combines final-answer correctness with an optional additional reward term for world-module usage.
\begin{equation}
\label{eq:ewm_reward}
r_{\mathcal{M}}(\mathcal{T},y^\star)
=
r(\hat{y},y^\star)
+
r_{\mathcal{W}}(\mathcal{T}).
\end{equation}
Here, \(r_{\mathcal{W}}(\mathcal{T})\) denotes the optional implementation-dependent reward for world-module calling behavior in the trace. We address this reward in Section~\ref{sec:selective_thought_experiments}.

A GRPO-style implementation \cite{shao2024deepseekmath} can optimise this reward with a clipped surrogate. For each training pair \((x,y^\star)\sim\mathcal{D}\), sample \(\mathcal{G}\) complete EWM rollouts \(\tau_{1:\mathcal{G}}=\{\tau_i\}_{i=1}^{\mathcal{G}}\) using the frozen old reasoner \(\pi_{\mathrm{old}}\) with access to \(\mathcal{W}\).

Let \(r_i=r_\mathcal{M}(\tau_i,y^\star)\). The group-relative advantage of rollout \(i\) is
\[
A_i
=
\frac{r_i-\bar{r}}{s_r+\epsilon_{\mathrm{adv}}},
\]
where \(\bar{r}\) and \(s_r\) are the mean and standard deviation of the \(r_i\) within \(\tau_{1:\mathcal{G}}\).

Let \(\tau_i=(z_{i1},\ldots,z_{iL_i})\) denote the serialized trajectory, where \(L_i\) is its length and \(z_{it}\) is the token at position \(t\). Also let \(\mathbbm{1}_{it}=1\) for policy-generated tokens and \(\mathbbm{1}_{it}=0\) for returned rollout observations. Define \(L_i^g=\sum_{t=1}^{L_i}\mathbbm{1}_{it}\), and, for generated tokens, 
\[
\rho_{it}
=
\frac{
\pi_\theta(z_{it}\mid \tau_{i,<t})
}{
\pi_{\mathrm{old}}(z_{it}\mid \tau_{i,<t})
}.
\]
Writing \(\operatorname{clip}_\epsilon(r)=\operatorname{clip}(r,1-\epsilon,1+\epsilon)\), the EWM training objective is
\begin{equation}
\label{eq:ewm_objective}
\begin{aligned}
J_E(\theta)
=
\mathbb{E}_{x,\,\tau_{1:\mathcal{G}}}
\Bigg[
&\frac{1}{\mathcal{G}}\sum_{i=1}^{\mathcal{G}}
\frac{1}{L_i^g}\sum_{t=1}^{L_i}\mathbbm{1}_{it}
\\[-0.25em]
&\quad
\min\!\left(
\rho_{it}A_i,\,
\operatorname{clip}_\epsilon(\rho_{it})A_i
\right) 
\\[-0.25em]
&\quad
-\beta D_{\mathrm{KL}}(\pi_\theta\|\pi_{\mathrm{ref}})
\Bigg].
\end{aligned}
\end{equation}

Here, \(\pi_{\mathrm{ref}}\) is the original pre-RL LLM used as a reference policy, and \(\beta\geq0\) controls the KL penalty that discourages drift from it.

\textit{Feasibility.} Prior work suggests that language models can acquire structured representations of space, time, colour, and other real-world variables from language alone \citep{gurnee2024language,huh2024platonic}. The LLM already contains much of the world knowledge needed to query a world-module effectively. Thus, the remaining task is not necessarily fresh pretraining from scratch, but targeted post-training.  Furthermore, autoregressive reasoning and video rollouts share a forward temporal structure. A chain of thought advances token by token, while a visualised scene advances frame by frame. The unidirectional nature of autoregression is surprisingly robust in practice \citep{nwadike-etal-2025-library}. It is also reflected in models that combine autoregressive prediction with latent diffusion sampling \citep{parkerholder2024genie2}.

%%%%%%%%%%%%%%%%%%%%%%%%%%%%%%%%%%%%%%%%%%%%%%%%%%%%%%%%%%%%%%%%%%%%%%%%%%%%%%%%%%%%%%%%%%% 

\subsubsection{\textit{Selective Thought Experiments}}
\label{sec:selective_thought_experiments} 
 
The \(r_{\mathcal{W}}\) term is intended to encourage selective world-module use. If \(r_{\mathcal{W}}(\mathcal{T})=0\), then selectivity can be learned solely through the final-answer reward, as in \citet{jin2025searchr1} and \citet{qian2025toolrl}. Emerging evidence, however, suggests that shaping rewards beyond final-answer correctness can encourage more specific behaviours, from improved reasoning to more efficient tool use \citep{guo2025deepseekr1,chen2026learning,wang2025actingless}. For example, one may set \(r_{\mathcal{W}}(\mathcal{T})=-\lambda M(\mathcal{T})/B\), where \(M(\mathcal{T})\) counts world-module calls, \(B\) is a call budget, and \(\lambda \ge 0\) controls the penalty for excessive calling. 

Upon choice of \(r_{\mathcal{W}}\), Algorithm~\ref{alg:ewm_training} summarises how the resulting reward \(r_\mathcal{M}\) may be utilised during training. Optimising \(J_E\) in Eq.~\ref{eq:ewm_objective} converts this trace-level reward into group-relative advantages over complete EWM trajectories. Traces where a visual thought experiment proves useful enough to justify its world-module usage receive higher advantages, while unnecessary or unhelpful calls receive lower advantages. The reasoner therefore learns when to query \(\mathcal{W}\) and how to use the returned rollout.
 
\begin{algorithm}[H]
\caption{Proposed EWM RLVR protocol}
\label{alg:ewm_training}
\algmid
\begin{algorithmic}[1]
\RaggedRight
\Require Dataset \(\mathcal{D}\), reasoner \(\pi_\theta\), reference policy \(\pi_{\mathrm{ref}}\), world-module \(\mathcal{W}\), group size \(\mathcal{G}\)
\For{each GRPO update round}
    \State Set frozen old policy \(\pi_{\mathrm{old}}\gets\pi_\theta\)
    \State Sample training batch from \(\mathcal{D}\)
    \For{each \((x,y^\star)\) in the batch}
        \State Sample \(\mathcal{G}\) trajectories \(\tau_{1:\mathcal{G}}\)
        \Statex \hspace{3.0em}using \(\pi_{\mathrm{old}}\) with access to \(\mathcal{W}\)
        \For{each trajectory \(\tau_i\)}
            \State Generate \ewmThinkOpen{}, \ewmToolOpen{},
            \Statex \hspace{4.2em} or \ewmAnswerOpen{} segments
            \If{the world-module is invoked}
                \State Call \(\mathcal{W}\) with the query 
                \Statex \hspace{5.8em} in \ewmToolOpen{} 
                \State Append rollout as a 
                \Statex \hspace{5.8em} \ewmRolloutOpen{} segment
            \EndIf
            \State Compute reward \(r_i=r_\mathcal{M}(\tau_i,y^\star)\)
        \EndFor
        \State Compute group-relative advantages \(A_i\)
    \EndFor
    \State Compute importance weights \(\rho_{it}\)
    \Statex \hspace{1.3em} over reasoner-generated tokens
    \State Mask returned visual-rollout observations \Statex \hspace{1.3em} from the policy loss
    \State Update \(\pi_\theta\) by maximising \(J_E\) in Eq.~\ref{eq:ewm_objective}
\EndFor
\end{algorithmic}
\end{algorithm}

%%%%%%%%%%%%%%%%%%%%%%%%%%%%%%%%%%%%%%%%%%%%%%%%%%%%%%%%%%%%%%%%%%%%%%%%%%%%%%%%%%%%%%%%%%% 

\subsection{\textit{The Latent Variant of EWMs}}

Continuous-space autoregression has already been demonstrated for text-based reasoning \citep{hao2025coconut}, and has been shown to have expressivity advantages \citep{zhu2025superposition}. In the latent variant of Einstein World Models, the visual-temporal rollout generated by the world-module is not directly used by the LLM reasoner. Instead, the latent variables corresponding to that rollout are used as the intermediate reasoning representation. In other words, image or video rollouts provide the supervision signal for the world-module during training, while at inference the corresponding latent representations from the world module are passed directly into the Einstein reasoner for subsequent autoregressive computation. 

The baseline discrete EWM variant first decodes the explicit rollout into an image or video, which is then re-encoded for use by the reasoner. This has the advantage of compatibility with existing vision encoders, while keeping the rollout directly inspectable, but requires the additional generation and encoding of the image or video during inference. However, the latent variant removes this additional inference computation, while providing a continuous interface in which linguistic and visual-temporal information can be represented within the same autoregressive reasoning process. 

The corresponding trade-off is that the reasoner must be trained to consume these latent semantic representations directly, rather than relying on an off-the-shelf vision encoder. At inference time, the latent rollout can still be decoded into an explicit visualisation when requested, since the world-module is trained to generate the corresponding visual-temporal rollout.

\label{sec:latent_variant}

%%%%%%%%%%%%%%%%%%%%%%%%%%%%%%%%%%%%%%%%%%%%%%%%%%%%%%%%%%%%%%%%%%%%%%%%%%%%%%%%%%%%%%%%%%%%%%%%%%%%%%%%%%%%%%%%%%%%%%%%%%%%%%%%%%%%%%%%%%%%%%%%%%%%%%%%%%%%%%%%%%%%%%%%%%%%%%%%%%%%%%%%%%%%%%%%%%%%%%%%%%%%%%%%%%%%%%%%%%%%%%%%%%%%%%%%

\section{\textit{World-Mod\underline{ule} Selection}}
\label{sec:world_module_selection}

Since RLVR optimises the reasoner's use of the world-module rather than the world-module itself, the question becomes how to select a module that produces useful visual-temporal thought experiments.

\subsection{Architecture}
\label{sec:architecture}

World-modules may come in several forms. We modify recent informal taxonomies \citep{li2026functionaltaxonomy} of world \textit{models} to distinguish candidate world-\textit{modules}:

\begin{enumerate}[leftmargin=1.2em, itemsep=0.5em, parsep=0em, topsep=0.5em, partopsep=0em]

    \item \textit{Renderers}: return observations, such as images or video frames. In \textit{EWMs}, renderers are the default world-modules. Text-to-video generators, or image-to-video generators used within a text-to-image pipeline, are of particular interest. This includes (but is not limited to) diffusion and flow-matching video models, provided their rollouts expose information that can support later reasoning \citep{kong2024hunyuanvideo,wan2025wan,hacohen2024ltxvideo}.

    \item \textit{Simulators}: allow the reasoner to intervene in a visualised world and observe what follows. Interactive world models such as Genie-style systems provide one example of this interface \citep{bruce2024genie,parkerholder2024genie2}, and benchmarks such as WBench suggest that the coherence of such interactive rollouts can be measured \citep{ying2026wbench}. 
    However, in practice, repeated renderer calls may play the role of a simulator, making a \textit{renderer} (aforementioned) the primary world-module of interest. In particular, the LLM can inspect one visualised consequence, revise its hypothesis, and then request another rollout from a modified condition or counterfactual premise. Simulators are therefore useful when a thought experiment requires explicit intervention, but they are not required for the central \textit{Einstein World Model} mechanism.

    \item \textit{Planners}: In \textit{Einstein World Models}, planning remains the role of the LLM reasoner, since the aim is visual-temporal reasoning over thought experiment sequences, rather than embodied robotic action.

\end{enumerate}

\subsection{World-Module Quality}
\label{sec:quality}
Einstein's visual thought experiments were disciplined by strong physical intuition. A generated video rollout may therefore only be as useful as the physical intuitions sustained by its underlying video model.

Diffusion models remain strong candidates because their intuitive-physics quality can be measured through human-verifiable likelihood estimates derived from the denoising objective of the diffusion model. \citet{yuan2025likephys} use precisely this technique to find substantial differences between video diffusion models, with stronger performance from recent systems. Some video generators remain unreliable physical simulators \citep{bansal2025videophy,zhang2025morpheus,motamed2026generative}. However, not all diffusion models are equal, and the frontier is fast-improving. Recent physics-aware generators further suggest that explicit dynamical priors can improve physical consistency and controllability \citep{yuan2026newtongen}.

Einstein World Models also require rollout faithfulness. A faithful rollout exposes information that the reasoner actually uses and that the final answer depends on. An unfaithful rollout, much like in traditional chain-of-thought, may not reveal the computation that produced the answer \citep{turpin2023language,lanham2023measuring}. The response is not to abandon the need for video rollout traces, but to evaluate and improve their faithfulness \citep{nwadike2026measuringaireasoningguide}. 

\subsection{\textit{Ensembling}}

\label{sec:ensembling} 

Einstein World Models already allow repeated world-module calls within a single reasoning trace. Here, \textit{ensembling} extends beyond repeated calls to a single module to describe the comparison of visual hypotheses generated by different world-modules, much as different humans may visualise the same problem differently. One module may favour visual realism, another physical consistency, and another temporal continuity. Because rollouts are externalised, these assumptions can be compared rather than left hidden. The problem is therefore not only to select a strong world-module, but to select modules whose inductive biases are usefully different. 

In \textit{ensembling}, different LLM reasoners, attached to different world-modules, exchange rollouts and critiques before answering. Each rollout proposes a different interpretation of the scene, and disagreement reveals what must be inspected next.

%%%%%%%%%%%%%%%%%%%%%%%%%%%%%%%%%%%%%%%%%%%%%%%%%%%%%%%%%%%%%%%%%%%%%%%%%%%%%%%%%%%%%%%%%%%%%%%%%%%%%%%%%%%%%%%%%%%%%%%%%%%%%%%%%%%%%%%%%%%%%%%%%%%%%%%%%%%%%%%%%%%%%%%%%%%%%%%%%%%%%%%%%%%%%%%%%%%%%%%%%%%%%%%%%%%%%%%%%%%%%%%%%%%%%%%%

% \cite{merrill2023parallelism,merrill2023expressive} 
\section{Related Work}
\label{sec:related_work}  

Chain-of-thought \cite{wei2022chain} made intermediate reasoning visible, but made it visible only as language. Many commonsense questions depend on variables that text traces represent poorly, including object identity, containment, contact, heat, motion, and material state. Humans often reason about such variables through visualisation. Einstein World Models seek to provide an analogous capability for language models by treating visualised visual episodes as intermediate reasoning artifacts. 

Whiteboard-of-Thought \cite{menon2024whiteboard} is a notable procedural predecessor for Einstein World Models. It gives a multimodal model a visual scratchpad, asks it to draw intermediate reasoning steps as an image, often through code, and then feeds that image back into the model for final reasoning. Einstein World Models preserve this feedback loop, but change the artifact from a static drawing to a short video rollout that serves as a visual play-through of a hypothesis.

\citet{hu2024visualsketchpad} similarly give multimodal LMs an external visual sketchpad for intermediate reasoning. However, the resulting artifacts are mainly static visual annotations (auxiliary lines, bounding boxes, segmentation masks, etc.) produced through programmatic tools, rather than visual-temporal rollouts.
 
\citet{wu2024mindseye} propose Visualization-of-Thought (VoT), a related method for spatial reasoning in LLMs. However, VoT focuses on controlled 2D grid-world tasks rather than real-world visual-temporal thought experiments. Its visualisations remain text-form grids or maps, not separate visual-temporal rollouts. Thus, VoT is best understood as a prompting strategy for symbolic state tracking, whereas the aim of Einstein World Models is to let a reasoner call a world-module and incorporate the resulting rollout into its trace.

In other work, \citet{chern2025generatedimages} showed that when generating an image from a prompt, intermediate visual subgoals can guide the model toward a better final image. However, their work focused on the objective of image generation as the end goal, rather than enhanced reasoning as the end goal (the latter being merely facilitated by visualisation as a means to an end).

\citet{tong2025thinkingvideo} study whether video generation models can reason by producing answer-bearing videos. Einstein World Models pursue a different research objective. Rather than asking whether the video generator can serve as the reasoner, EWMs ask whether an LLM can use video generation as a thought experiment tool. In other words, rather than replacing the LLM with a video generator, we ask how frontier LLMs can decide when visualisation is useful, selectively invoke the world-module, and integrate the resulting rollout back into reasoning.

\citet{yang2025mindjourney} and \citet{yu2026when} explore the importance of visualisation for 3D spatial reasoning in VQA-style settings, where the model begins from an observed image and generates additional views to answer questions about that scene. This differs from the setting of interest here, where thought experiments are used to support text-based reasoning problems and the model must decide, as part of its own reasoning process, when visual-temporal rollouts are useful.

We discuss the relationship between Einstein World Models and contemporary agentic LLM systems in the Supplementary Notes, Part ~\ref{appendix:relation_to_agents}.

%%%%%%%%%%%%%%%%%%%%%%%%%%%%%%%%%%%%%%%%%%%%%%%%%%%%%%%%%%%%%%%%%%%%%%%%%%%%%%%%%%%%%%%%%%%%%%%%%%%%%%%%%%%%%%%%%%%%%%%%%%%%%%%%%%%%%%%%%%%%%%%%%%%%%%%%%%%%%%%%%%%%%%%%%%%%%%%%%%%%%%%%%%%%%%%%%%%%%%%%%%%%%%%%%%%%%%%%%%%%%%%%%%%%%%%%
\section{Future Work: A Call for Datasets}
\label{sec:call_for_datasets}
\vspace{0.5em}
A central bottleneck for Einstein World Models is \textit{data}. Few existing datasets explicitly target the behaviour Einstein World Models require. The missing setting is neither ordinary text reasoning, nor visual question answering over an already provided image. It is a setting in which an LLM is fully capable of taking in text alone as input (even without the aid of visual inputs, when requested), and can perform a visual thought experiment in its reasoning trace before outputting final answer tokens. This makes dataset construction the immediate experimental bottleneck for Einstein World Models, since the core learning problem still lacks a suitable benchmark.

SimpleBench is one of the rare public datasets pointing in this direction. However, the full dataset contains only a little over 200 questions, while its public release exposes only 10, making it useful as an illustration, rather than as a training corpus. The questions in SimpleBench are short and text-only, yet difficult, because answering them correctly often depends on performing a thought experiment about how a described scene unfolds. Consider the following example:\begin{datasetexample}{SimpleBench  \titlecite{philip2024simplebench} \\Test Sample \#2} 
\vspace{0.1em}\noindent\textbf{Question:} A juggler throws a solid blue ball a meter in the air and then a solid purple ball (of the same size) two meters in the air. She then climbs to the top of a tall ladder carefully, balancing a yellow balloon on her head. Where is the purple ball most likely now, in relation to the blue ball?

A. at the same height as the blue ball

B. at the same height as the yellow balloon

C. inside the blue ball

D. above the yellow balloon

E. below the blue ball

F. above the blue ball

\vspace{0.5em}\noindent\textbf{Correct Answer:} A. 
\end{datasetexample}
 
What makes this question difficult for LLMs is that it requires visualising the scene, and realising that, although the purple ball is thrown higher than the blue ball, enough time passes for both solid balls to fall back down before any juggler could possibly finish climbing a ladder while carefully balancing a balloon on their head. A text-only LLM may over-formalise the prompt and conclude that the answer depends on unspecified variables such as launch velocity or climbing time. However, a one- or two-metre throw lasts only moments, while carefully climbing a tall ladder takes long enough for ordinary solid balls to land. An EWM rollout would externalise such a missing visual-temporal computation, allowing it to become part of the reasoning trace.

Existing physical reasoning datasets are valuable, but many begin with the relevant scene already available. Some focus on two-dimensional puzzle images \citep{bakhtin2019phyre}. Others ask models to explain, predict, or judge physical events in supplied video clips \citep{yi2020clevrer,bear2021physion,riochet2018intphys,bordes2025intphys2}. Still, others focus on the physical fidelity of video generators. Even then, the visual scene is supplied in advance, either as initial frames to complete, or as rendered samples to score \citep{upadhyay2026worldbench,yuan2025likephys}. By contrast, Einstein World Models need datasets where the problem begins as language, and the model must decide, for itself, whether to generate a visualisation.

We therefore make an open call for datasets in this setting. Ideally, such datasets should contain both problems that benefit from visual thought experiments, and problems that do not, allowing Einstein World Models to learn not only how to visualise, but also when not to.

%%%%%%%%%%%%%%%%%%%%%%%%%%%%%%%%%%%%%%%%%%%%%%%%%%%%%%%%%%%%%%%%%%%%%%%%%%%%%%%%%%%%%%%%%%%%%%%%%%%%%%%%%%%%%%%%%%%%%%%%%%%%%%%%%%%%%%%%%%%%%%%%%%%%%%%%%%%%%%%%%%%%%%%%%%%%%%%%%%%%%%%%%%%%%%%%%%%%%%%%%%%%%%%%%%%%%%%%%%%%%%%%%%%%%%%%
% \vspace{1.8em}
\section*{Epilogue}
\label{sec:epilogue}
\vspace{0.6em}
 
\textit{This paper proposed Einstein World Models, for treating visual thought experiments as a tool-use behaviour in LLM reasoning. The central idea is that some questions require a model to visualise how a described scene unfolds, and that this may be poorly supported by language alone. EWMs keep the LLM as the reasoner, but allow it to call a world-module, render a short visual-temporal rollout, and return that rollout to the reasoning trace before answering. A video rollout is not assumed to be a perfect simulation, but rather a visual hypothesis about how a described situation may unfold. Its distinctive usefulness comes from being externalised, making the model's reasoning process available for inspection and debugging. The path forward thus requires both better world-module curation, as well as better datasets to allow for training LLMs on when and how thought experiments should be invoked. In this sense, Einstein World Models point toward language models that reason not only through words and symbols, but through externalised visual walk-throughs. }

% \newpage

%%%%%%%%%%%%%%%%%%%%%%%%%%%%%% 
\appendix
\section*{Supplementary Notes}

\section{\textit{Supervised Finetuning}}
\label{appendix:sft}
Before reinforcement learning, the Einstein reasoner may be warm-started with supervised fine-tuning on valid EWM trace formats using a standard cross-entropy loss. This stage teaches the syntax and role structure of EWM reasoning traces, including ordinary language reasoning, world-module query segments, returned visual-rollout observations, and final-answer segments. For example, traces may use \ewmThinkOpen{} for language reasoning, \ewmToolOpen{} for world-module queries, \ewmRolloutOpen{} for returned observations, and \ewmAnswerOpen{} for final answers.

Let
\[
\mathcal{D}_{\mathrm{SFT}}
=
\{(x_i,\mathcal{T}_i^\star)\}_{i}
\]
be a supervised dataset of input problems and target EWM traces. Each trace \(\mathcal{T}_i^\star\) contains both reasoner-generated segments and observation segments returned by the world-module. The model is trained with standard next-token cross-entropy, but only on tokens generated by the reasoner. Returned visual rollouts are observations, not policy actions, and are therefore masked from the supervised loss.

Let \(z_t^\star\) denote the target token at position \(t\) in the serialized trace \(\mathcal{T}^\star\). Let \(\mathbbm{1}_t=1\) if \(z_t^\star\) is a target token that the reasoner is expected to produce, and let \(\mathbbm{1}_t=0\) if \(z_t^\star\) is part of a returned rollout observation. The masked supervised fine-tuning loss is:
\[
\begin{aligned}
\mathcal{L}_{\mathrm{SFT}}(\theta)
&=\\
& \hspace{-3.7em}
 -\mathop{\mathbb{E}}\limits_{\substack{(x,\mathcal{T}^\star)\sim
 \mathcal{D}_{\mathrm{SFT}}}}
\left(\dfrac{1}{\sum_t \mathbbm{1}_t}
\sum_t
\mathbbm{1}_t
\log \pi_\theta
\bigl(z_t^\star \mid \mathcal{T}_{<t}^\star\bigr)
\right)\hspace{-0.2em}.
\end{aligned}
\]

Thus, supervised fine-tuning prepares the Einstein reasoner \(\pi_\theta\) for RLVR by teaching it the format of valid EWM traces, while RLVR teaches when and why such traces should contain visual thought experiments. Crucially, \(\mathcal{D}_{\mathrm{SFT}}\) should include both call and no-call traces, so that the model learns the world-module interface without learning to invoke \(\mathcal{W}\) by default.

%%%%%%%%%%%%%%%%%%%%%%%%%%%%%%%%%%%%%%%%%%%%%%%%%%%%%%%%%%%%%%%%%%%%%%%%%%%%%%%%%%%%%%%%%%% 

\section{\textit{Relation to Agentic Systems}}
\label{appendix:relation_to_agents}

Modern LLM agents are commonly understood as language models augmented with additional machinery for tool use, planning, environment interaction, or action execution on behalf of a user \citep{yao2023react,shen2023hugginggpt}. However, although an EWM can be incorporated into an agentic system, it is not necessarily an agent in itself. Such systems may qualify as Einstein World Models when they use a world-module to generate visual-temporal rollouts as intermediate reasoning artifacts, but not merely by virtue of being agents. Existing agentic systems can call tools, but they do not by default perform externalised visual thought experiments in the sense proposed here. Conversely, an Einstein World Model need not have the broader infrastructure of a general-purpose agent for executing user-facing actions, such as managing software repositories or querying databases. It may be implemented as a language reasoner that calls a world-module only to imagine a described scene, inspect the resulting rollout, and incorporate it into its reasoning trace. In this sense, EWMs describe a reasoning capability that can either stand alone, or be added to agentic systems.

%%%%%%%%%%%%%%%%%%%%%%%%%%%%%%%%%%%%%%%%%%%%%%%%%%%%%%%%%%%%%%%%%%%%%%%%%%%%%%%%%%%%%%%%%%% 

\section{\textit{Relation to Other World Models}}
\label{appendix:relation_to_wms}

The term \textit{world model} has a long history in AI, especially in model-based reinforcement learning, where learned models of environment dynamics support planning, control, and interaction with an environment \citep{sutton1991dyna,ha2018recurrent}. In this lineage, a world model helps an agent anticipate possible futures before acting in the external world. More recent proposals similarly treat world models as central components of autonomous systems that learn to predict, reason, and plan over future states \citep{lecun2022path}.

Einstein World Models shift the emphasis from the notion of a world \textit{model} to that of a world-\textit{module}. An EWM is not itself a learned dynamics model, simulator, or video generator. It is a reasoning system in which the LLM remains the reasoner and calls the world-module when a visual-temporal thought experiment may support its reasoning trace. The world-module is the component most closely aligned with what is traditionally the predictive component of a world model. It need not be a full 3D simulator, and is treated only as a component of a reasoning system. Its role in an EWM is to supply an inspectable rollout rather than to replace the LLM's reasoning process.

This distinction also shifts what must be learned. Whereas many world-model architectures focus on modelling environmental dynamics, EWMs focus on how a reasoner should selectively invoke visualisation to facilitate reasoning.  
%%%%%%%%%%%%%%%%%%%%%%%% 

\bibliography{references}

%%%%%%%%%%%%%%%%%%%%%%%%%%%%%%%%%%%%%%%%%%%%%%%%%%%%%%%%%%%%%%%%%%%%%%%%%%%%%%%%%%%%%%%%%%%%%%%%%%%%%%%%%%%%%%%%%%%%

% \newpage
% \appendix

% \include{arr_format/supplementary} 
%%%%%%%%%%%%%%%%%%%%%%%%%%%%%%%%%%%%%%%%%%%%%%%%%%%%%%%%%%%%%%%%%%%%%%%%%%%%%%%%%%%%%%%%%%%%%%%%%%%%%%%%%%%%%%%%%%%%

% \include{old_stuff}
 
\end{document}